%% file: root_conf.tex
\documentclass[letterpaper, 10 pt, conference]{ieeeconf}  % Comment this line out if you need a4paper
\IEEEoverridecommandlockouts

\usepackage[top=57pt,bottom=43pt,left=48pt,right=48pt]{geometry}
\usepackage{cite}
\usepackage{booktabs} % For prettier tables
\usepackage{graphicx}
\usepackage{xcolor}
\usepackage{subcaption}
\usepackage{graphicx}
\usepackage{url}
\usepackage{hyperref}
\usepackage{comment}

\usepackage{subfiles}
\usepackage{amsmath,amssymb,amsfonts,mathtools, dsfont, mathrsfs}
\usepackage{algorithm,algpseudocode}
\usepackage{bm, multirow, array, xfrac}
\newtheorem{theorem}{Theorem}
\newtheorem{remark}{Remark}

\newtheorem{definition}{Definition}

\newtheorem{assumption}{Assumption}
\usepackage{color}

\def\BibTeX{{\rm B\kern-.05em{\sc i\kern-.025em b}\kern-.08em
    T\kern-.1667em\lower.7ex\hbox{E}\kern-.125emX}}
\begin{document}

\title{\bf Floquet-Parameterized Koopman-MPC for Tracking Periodic Orbits}
\author{ 
Kartik A. Pant$^{1}$, Vishnu Vijay$^{1}$, Yufei Zhu$^{2}$, Xintong Deng$^{2}$, Taewoo Nam$^{2}$  and Inseok Hwang$^{1}$%
\thanks{$^{1}$The authors are with the School of Aeronautics and Astronautics, Purdue University,
West Lafayette, IN 47906. Email: \mbox{(kpant, vvijay, ihwang@purdue.edu)}}
\thanks{$^{2}$The authors are with the Toyota Research Institute of North America,
Ann Arbor, MI 48106. Email: \mbox{(yufei.zhu, xintong.deng, taewoo.nam@toyota.com)}}}
\maketitle
\thispagestyle{empty}

% OUR SUBFILES
\subfile{sections/0_abstract}
\section{Introduction}
\label{sec:introduction}
\subfile{sections/1_introduction}
\section{Preliminaries}
\label{sec:preliminaries}
\subfile{sections/2_preliminaries}
\section{Problem Formulation}
\label{sec:prob_form}
\subfile{sections/3_prob_form}
\section{Floquet-Parameterized Koopman-MPC}
\label{sec:main_res}
\subfile{sections/4_main_results}
\section{Numerical Simulations}
\label{sec:sim}
\subfile{sections/5_numerical_example}
\section{Conclusion}
\label{sec:conc}
\subfile{sections/6_conclusion}
\bibliographystyle{ieeetr}
\bibliography{IEEEabrv,refs} 
%\subfile{sections/7_appendix}
\end{document}

%% file: sections/0_abstract.tex
\begin{abstract} 
Designing real-time control for nonlinear systems to track periodic orbits can be challenging, especially for complex and coupled dynamical systems (e.g., tethered kites), where the true model parameters are hard to obtain. 
To address this, we propose a physics-informed Koopman learning approach that incorporates periodic-orbit stability directly into the learned linear predictor. The approach combines the underlying periodic structure of the dynamics along with periodic state-input trajectory data to learn physically consistent linear predictors (in lifted/observable space). We leverage Floquet theory, which studies the stability of periodic orbits in nonlinear systems, to obtain the periodic stability information of a given orbit via a monodromy matrix (i.e., the state transition matrix over an entire period).
Subsequently, we impose open-loop monodromy constraints to learn Floquet-parameterized linear predictors using a least-squares Extended Dynamic Mode Decomposition (EDMD) formulation, thereby constraining the search space and aiding in Koopman learning. We then utilize these linear predictors to design a model predictive control (MPC) law for tracking periodic references. Finally, we evaluate the designed framework on a cartpole system under periodic forcing and compare it against a purely data-driven baseline. 
\end{abstract}

\begin{keywords}
Koopman operator, Floquet theory, periodic orbits, data-driven method, and model predictive control.
\end{keywords}

%% file: sections/1_introduction.tex
Many complex engineering systems exhibit periodic behavior during operation, either naturally or in response to a forcing input (e.g., walking in humanoid robots, tethered kites looping in a figure-of-eight path, etc.). In many cases, disturbances and environmental effects push these systems far from their reference, leading to sustained periodic oscillations that can build up and cause catastrophic failure (e.g., a humanoid robot falling to the ground or kites spinning violently). Thus, tracking controllers must stabilize trajectories despite large deviations from the reference, enabling resilience to unforeseen disturbances.

Nonlinear model predictive control (NMPC) has garnered significant attention in recent years for enabling optimization-based control with theoretical guarantees on stability and constraint satisfaction. However, the biggest challenge in deploying NMPC-based tracking controllers is model complexity and its impact on real-time performance. In particular, for complex, coupled dynamical systems, computational overhead may prevent real-time NMPC implementation. To address these challenges, attention has rapidly shifted toward Koopman operator theory, which comes with an auspicious promise of global representation of nonlinear systems by a linear operator -- one that could transform the deployment of mathematically interpretable learning-based controllers for real-world applications.  

A significant body of literature explores data-driven methods for learning the Koopman operator \cite{Brunton2022Modern, mauroy2020koopman} and its applications in control \cite{strasser2026overview, iacob2024koopman, korda2018linear}. This spurred increased interest in purely data-driven approaches that learn Koopman models directly from data, such as extended dynamic mode decomposition (EDMD)\cite{Brunton2022Modern}. Although powerful, these methods tend to overfit limited data rather than capture the underlying physics when the lifting functions are not chosen appropriately. Recent works have examined enforcing key dynamic characteristics while learning Koopman models, such as stability \cite{dahdah2022system, mamakoukas2023learning}, dissipativity \cite{hara2020learning, xu2025learning}, and sector boundedness \cite{xu2025learning}, to make model learning sample-efficient and physically consistent. For periodic systems, \cite{mezic2016koopman} extends Koopman theory by leveraging ideas from Floquet theory \cite{Floquet1883}, resulting in phase-parameterized linear Koopman models; however, the results are limited to autonomous systems without any control inputs. 

For control design, the seminal work by Korda et al. \cite{korda2018linear} established Koopman operators, particularly EDMD, to learn linear and bilinear lifted models and subsequently developed quadratic programs for MPC. Various extensions of Koopman-MPC for regulation and tracking control, along with stability guarantees, have been proposed \cite{wang2022data, de2024koopman,schimperna2026stability}. Most of the above works focus on general or control-affine nonlinear systems, and very few have explored using Koopman theory for control applications of periodic systems. 

To this end, we propose a constrained EDMD-based Koopman-MPC formulation for tracking periodic orbits. We first derive the relationship between the monodromy matrix (i.e., the state transition matrix over an entire period)  of the original system (along a periodic orbit) and that of the lifted system. Subsequently, we use the derived monodromy relation to learn a linear time-periodic model from state-input trajectory data using a constrained EDMD formulation. Finally, we use the learned lifted model, which captures/preserves the phase-based cyclic structure, to formulate the MPC optimization as a quadratic program in the lifted space for tracking periodic orbits.  

The rest of this paper is organized as follows. Section \ref{sec:preliminaries} presents the preliminaries on Floquet theory for nonlinear periodic systems. Section \ref{sec:prob_form} formulates the tracking control problem for periodic systems with model mismatch. In Section \ref{sec:main_res}, we present the relationship between the periodic stability of the original system and that of the learned system and utilize it in a constrained EDMD formulation to learn a lifted model. We then present a Koopman MPC formulation for control design to track periodic orbits using the learned model. Section \ref{sec:sim} provides an illustrative numerical example of our proposed approach. Section \ref{sec:conc} concludes the paper.

% \textbf{Notations:}

%% file: sections/2_preliminaries.tex
In this section, we present preliminaries on Floquet theory for nonlinear time-periodic systems.

\subsection{Floquet Theory: Stability of Periodic Solutions}
Consider a non-autonomous $T-$periodic discrete-time nonlinear system governed by
\begin{equation}
\label{eq:koop_nonlinear_periodic}
    x_{k+1} = f(x, k), \quad f(\cdot, k+T) = f(\cdot, k),
\end{equation}
where $k\in \mathbb{Z}$ is the time-index, $x_k\in \mathcal{X} \subseteq \mathbb{R}^{n_x}$ is the state trajectory starting at $x_{k_0} = x_0$, $f:\mathcal{X}\times\mathbb{Z} \rightarrow\mathcal{X}$ is a smooth map, and $T\in \mathbb{N}$ is the period. Let $\bar{x}$ denote the reference periodic orbit of the system in \eqref{eq:koop_nonlinear_periodic}. To analyze its stability properties, that is, whether the trajectories that are close to the orbit will converge to the orbit, stay in the vicinity of, or diverge from the orbit, we consider the dynamics of a perturbation $y_k = x_k - \bar{x}_k$ from the periodic solution with $y_{0} = y^0$. Substituting the dynamics in \eqref{eq:koop_nonlinear_periodic}, we get 
\begin{equation}
    y_{k+1} = f(x_k, k) - \bar{x}_{k+1}\coloneqq \zeta (k)y_k + \mathcal{O}(||{y}||^2),
\end{equation}
where $\zeta(k) = \frac{\partial f}{\partial\mathbf{x}}|_{\bar{x}_k, k}$ is the Jacobian evaluated along the trajectory of the periodic solution. The approximate linear time-varying (LTV) system can be written as
\begin{equation}
\label{eq:flo_approx_ltv}
    y_{k+1} = \zeta(k)y_k, \quad y_{k_0} = y_0. 
\end{equation}
The equilibrium of the LTV system in \eqref{eq:flo_approx_ltv} is at zero, corresponding to the periodic orbit of the original system. Thus, the stability analysis reduces to the stability of this equilibrium. The fundamental solution matrix (or the state transition matrix) $\mathbf{\Omega}(k)$ is the solution to 
\begin{equation}
    \mathbf{\Omega}(k+1) = \zeta(k)\mathbf{\Omega}(k); \quad \mathbf{\Omega}(0) = \mathbf{I}
\end{equation}
and any state can be obtained by $y_k = \mathbf{\Omega}(k) y_0$. The fundamental solution matrix evaluated after a complete period $T$, $\mathbf{\Omega}(T)$, is called a monodromy matrix, and its eigenvalues $\{\mu_l\}^n_{l=1}$ are called Floquet multipliers \cite{teschl2012ordinary}. Thus, the periodic solution is asymptotically stable if all Floquet multipliers lie strictly within the unit circle and unstable if any lies outside it. If multipliers lie on the unit circle with none outside, the solution is non-hyperbolic and requires further analysis to determine stability.

%% file: sections/3_prob_form.tex
% Consider a discrete-time $T-$periodic parametric nonlinear system governed by
% \begin{equation}
% \label{eq:nonlinear_dyn}
%     x_{k+1} = f_{p}(x_k, u_k, k), \quad f_{p}(\cdot, \cdot, k+T) = f_{p}(\cdot, \cdot, k). 
% \end{equation}
% where $x_k \in \mathbb{R}^{n_x}$ is the state, $u_k \in \mathbb{R}^{n_u}$ is the inputs, $T$ is the period and $p\in \mathcal{P} \subseteq \mathbb{R}^{n_p}$ is the parameters. 
% \begin{remark}
% The dynamics in \eqref{eq:nonlinear_dyn} is a generalization over a larger class of nonlinear $T-$periodic systems, which are periodic either due to autonomous systems on a periodic orbit (e.g., pendulums) or are under exogenous periodic forcing (e.g., humanoid robots). In the  case of exogenous forcing, $u_{k+T} = u_k$ forces the dynamics to satisfy $f_{p}(\cdot, u_{k+T}, k+T) = f_{p}(\cdot, u_k, k)$.
% \end{remark}
% Suppose $ \{\mathbf{\bar{x}}, \mathbf{\bar{u}}\}$ denote a periodic orbit of the nonlinear dynamical system in \eqref{eq:nonlinear_dyn}. These periodic trajectories can be found using Harmonic Balance methods (HBMs)\cite{} or obtained by solving optimal control problems (OCPs) \cite{}. In this work, we assume a mismatch between the actual model parameters and the nominal model. The goal is to design a controller that tracks the reference trajectory under model mismatch while satisfying input and state constraints. 

% \red{MAIN CHALLENGES in Control Design}

Consider a discrete-time $T-$periodic parametric nonlinear system governed by
\begin{equation}
    \label{eq:nonlinear_dyn}
    x_{k+1} = f_{p}(x_k, u_k, \sigma_k),  \ f_{p}(\cdot, k+T) = f_{p}(\cdot, k).
\end{equation}

where $x_k \in  \mathcal{X} \subseteq \mathbb{R}^{n_x}$ is the state, $u_k \in \mathcal{U} \subseteq\mathbb{R}^{n_u}$ is the input, $T$ is the period, $\sigma_{k+1} = (\sigma_k+1) \bmod T$ , with $\sigma_0 = 0$, and $p\in \mathcal{P} \subseteq \mathbb{R}^{n_p}$ are the parameters of the system dynamics. 
% The system is subject to the constraints
% \begin{equation}
% \label{eq:constraints}
%     x_k \in \mathcal{X} \subseteq \mathbb{R}^{n_x}, \qquad u_k \in \mathcal{U} \subset \mathbb{R}^{n_u},
% \end{equation}
The set $\mathcal{X}$ is closed, $\mathcal{U}$ is compact, and both are polytopic.
% \begin{remark}
% The dynamics in \eqref{eq:nonlinear_dyn} is a generalization over a larger class of nonlinear $T-$periodic systems, which are periodic either due to autonomous systems on a periodic orbit (e.g., pendulums) or are under exogenous periodic forcing (e.g., humanoid robots). In the case of exogenous forcing, $u_{k+T} = u_k$ forces the dynamics to satisfy $f_{p}(\cdot, u_{k+T}, k+T) = f_{p}(\cdot, u_k, k)$.
% \end{remark}

\begin{remark}
The system dynamics in \eqref{eq:nonlinear_dyn} is a generalization of both sources of periodicity, i.e., autonomous systems evolving on a periodic orbit (e.g., a pendulum), and systems driven by exogenous periodic forcing (e.g., tethered kites).
\end{remark}

\begin{definition}[Periodic orbit]
\label{def:orbit}
A pair of sequences $\{\bar{\mathbf{x}}, \bar{\mathbf{u}}\}$, with $\bar{\mathbf{x}} = (\bar{x}_0,\dots,\bar{x}_{T-1})$ and $\bar{\mathbf{u}} = (\bar{u}_0,\dots,\bar{u}_{T-1})$, is a $T$-periodic orbit of \eqref{eq:nonlinear_dyn} if
\begin{equation}
\label{eq:orbit}
    \bar{x}_{k+1} = f_{p}(\bar{x}_k,\bar{u}_k,\sigma_k), \quad \bar{x}_{k+T} = \bar{x}_k, \quad \bar{u}_{k+T} = \bar{u}_k,
\end{equation}
for all $k \in \mathbb{Z}_{\geq 0}$, and $(\bar{x}_k,\bar{u}_k) \in \mathrm{int}(\mathcal{X}) \times \mathrm{int}(\mathcal{U})$.
\end{definition}

These periodic trajectories (or orbits) can be found using Harmonic Balance methods (HBMs)\cite{krack2019harmonic} or obtained by solving optimal control problems (OCPs) \cite{zanon2016periodic}. Define  $\delta x_k$ and $ \delta u_k$ as the variational state and inputs and linearizing \eqref{eq:nonlinear_dyn} about the orbit yields 
%the linear $T$-periodic (LTP) variational dynamics
\begin{equation}
\label{eq:variational}
    \delta x_{k+1} = \Omega^p_{\sigma_k} \, \delta x_k + \Gamma^p_{\sigma_k} \, \delta u_k,
\end{equation}
with $\Omega^p_{\sigma_k} \coloneqq \tfrac{\partial f_{p}}{\partial x}\big|_{(\bar{x}_k,\bar{u}_k,\sigma_k)}$, $\Gamma^p_{\sigma_k} \coloneqq \tfrac{\partial f_{p}}{\partial u}\big|_{(\bar{x}_k,\bar{u}_k,\sigma_k)}$, and $\Omega^p_{k+T} = \Omega^p_k$, $\Gamma^p_{k+T} = \Gamma^p_k$. The associated \emph{parametric monodromy matrix} and \emph{Floquet multipliers} are %at phase $\sigma_k \in \{ 0,\dots T-1\}$  are
\begin{equation}
\label{eq:monodromy}
    M^x_{\sigma_k}(p) \coloneqq \Omega^p_{\sigma_k+T-1}\cdots\Omega^p_{\sigma_k}, \ \ \Lambda(p) \coloneqq \mathrm{spec}(M^x_{\sigma_k}(p)),
\end{equation}
where $\Lambda(p)$ is independent of $\sigma_k$ and $\mathrm{spec}(\cdot)$ denotes the set of eigenvalues (e.g., eigen spectrum) of a matrix, since the eigenvalues of the monodromy matrices and their multiplicities are independent of the phase \cite{bittanti2009periodic}. The pair $(M^x_{\sigma_k}(p), \Lambda(p))$ encodes the local stability and the multi-step growth behavior of \eqref{eq:nonlinear_dyn} along the orbit.

In this work, we assume a mismatch between the actual model parameters and the nominal model. The true system evolves under an unknown $p \in \mathcal{P}$, while we only have nominal parameters, $\hat{p} \in \mathcal{P} \ | \ \hat{p} \neq p$, available. We also have a finite set of measured trajectories, denoted as
\begin{equation}
\label{eq:data}
    \mathcal{D} = \big\{ (x^{(i)}_k, u^{(i)}_k, x^{(i)}_{k+1}, \sigma_k^{(i)}) \big\}_{i=1}^{D},
\end{equation}
generated by the true system. The goal is to design a controller that tracks the reference trajectory under model mismatch while satisfying input and state constraints. In particular, we aim to solve the following $T$-periodic tracking problem in a receding horizon manner:
\begin{equation}
\label{eq:ideal_ocp}
\begin{aligned}
    \min_{\mathbf{u}} \quad & \sum_{i=0}^{H-1} \ell\big(x_{k+i} - \bar{x}_{k+i},\, u_{k+i} - \bar{u}_{k+i}\big) \\
    \text{s.t.} \quad & x_{k+i+1} = f_{\hat{p}}(x_{k+i}, u_{k+i}, \sigma_{k+i}), \\
    & x_{k+i} \in \mathcal{X}, \quad u_{k+i} \in \mathcal{U}, \quad x_k = x(k),
\end{aligned}
\end{equation}
where $f_{\hat{p}}(\cdot)$ is the nominal model, $H$ is the prediction horizon and $\ell$ is a positive-definite stage cost. Concisely, the problem statement is as follows: 

\textbf{Problem Statement:} \emph{Given the data $\mathcal{D}$ in \eqref{eq:data}, and the nominal model $f_{\hat{p}}(\cdot)$, construct a lifted linear $T$-periodic predictor whose monodromy is consistent with that of the true system \eqref{eq:monodromy} by construction. Using the predictor, solve the receding horizon constrained optimization problem \eqref{eq:ideal_ocp}.}

%% file: sections/4_main_results.tex
In this section, we first show how the Koopman operator for a $T-$periodic controlled nonlinear system can be transformed into an autonomous operator with an extended state. We then present a phase-binned EDMD approach that preserves the periodic structure. Furthermore, we derive the relationship between the monodromy matrix of the true system and that of the lifted system. Consequently, we propose a constrained EDMD approach that uses this relation and embeds the periodic stability information of the reference orbit into the lifted model, constraining the search space and enabling physically consistent linear predictors. Finally, using the obtained linear predictors, we present the linear MPC formulation for tracking periodic orbits.  
\subsection{Koopman Operator for T-Periodic Controlled Systems}
\label{sec:koopman_periodic}
We first show that the Koopman operator for $T-$periodic nonlinear controlled systems can be transformed into an autonomous system. Hence, we can use methods such as EDMD to find finite-dimensional approximations. Furthermore, we can retain the theoretical guarantees associated with the finite-dimensional approximation \cite{korda2018linear}.  

We extend the notion of an extended state used in \cite{korda2018linear} by defining $\chi = \left[ x_k^\top, \bm{u}^\top, \sigma_k\right]^\top \in \mathbb{R}^{n_x} \times \ell(\mathcal{U}) \times\mathbb{Z}_T $, where $\bm u\coloneqq \{ u(i)\}_{i=0}^\infty\in \ell(\mathcal{U})$ with $\ell(\mathcal{U})$ denoting the space of all the sequences $\{ u(i)\}_{i=0}^\infty$. The dynamics of the extended state is $f(\chi) = \begin{bmatrix}f(x_k, \bm{u}(0))^\top & (\mathcal{S}\bm u)^\top &\sigma_{k+1}\end{bmatrix}^\top$, where $\mathcal{S}$ denotes the left shift operator, i.e., $\mathcal{S}\bm u(i) = \bm u(i+1)$, $\bm u(i)$ denote the $i^{th}$ element of the sequence $\bm u$ and $\sigma_{k+1} = (k+1)\ \text{mod}\ T$ is the phase shift. Then, we can define the Koopman operator associated with the dynamics of the extended state $\chi$, on the set of observables $\phi(\chi)$, as $(\mathcal{K}\phi)(\chi) \coloneqq \phi\circ f(\chi)$.

\subsection{Phase-binned EDMD and Block-cyclic System Structure}
In practice, we truncate the infinite-dimensional Koopman operator for downstream applications, e.g., control (see \cite{korda2018linear}). In this work, we leverage extended dynamic mode decomposition (EDMD), a data-driven algorithm that constructs the Koopman operator from state-input trajectory data. However, periodic systems retain a cyclic property across system phases. Conventional EDMD for control (EDMDc) loses this periodic structure. To address this, we use phase-binned EDMDc, where the key change is binning the trajectory data (and subsequently the lifting) based on the phase. %Define $\psi^{[\sigma_k]}(x) \coloneq [0,.., \psi(x),.., 0]^\top \in \mathbb{R}^{n_{\psi}T}$ and $\bm u^{[\sigma_k]} \coloneq [0,.., \bm u(0),.., 0]^\top \in \mathbb{R}^{n_uT}$ with slots filled based on phase $\sigma_k$, where $\psi(x) = [\psi_1(x), \dots, \psi_{n_\psi}(x)]^\top \in \mathbb{R}^{n_\psi}$ and $\bm u(0) \in \mathbb{R}^{n_u}$ denote the first element of the sequence $\bm u$. 
Define $\psi^{[\sigma_k]}(x) = e_{\sigma_k}\otimes\psi(x), \ \bm u^{[\sigma_k]} = e_{\sigma_k}\otimes \bm u(0)$ as the phase-binned lifting of the state and inputs, where $e_{\sigma_k} = \begin{bmatrix} \mathbb{I}(\sigma_k=1) & \cdots & \mathbb{I}(\sigma_k=T)\end{bmatrix}^{\top}$ and $\bm u(0) \in \mathbb{R}^{n_u}$ denotes the first element of the sequence $\bm u$. 
We then construct the observables in the following manner:
\begin{equation}
    \phi(x, \bm u, \sigma_k) = \begin{bsmallmatrix}\psi^{[\sigma_k]}(x) \\ \bm u^{[\sigma_k]}\end{bsmallmatrix} \in \mathbb{R}^{n_\phi},
\end{equation}
where $n_\phi = n_\psi T+n_uT$ with $n_\psi$ denoting the dimensions of the state observables $\psi(\cdot)$. Assuming that we obtain the phase-binned data (i.e., each data point contains the associated phase) in \eqref{eq:data} of the form $(\chi_j, \chi_j^{+}), j =\{1 \dots D\}$ satisfying $\chi^+ =f(\chi) $, the EDMD algorithm seeks to find $\mathcal{A} \in \mathbb{R}^{n_\phi\times n_\phi}$ minimizing, 
\begin{equation}
    \min_\mathcal{A}\sum_j^D ||\phi(\chi_j^{+}) - \mathcal{A}\phi(\chi_j)||^2_2.
\end{equation}
Note that the matrix $\mathcal{A}$ has a sparse block-cyclic internal structure where the first $n_\psi T$ rows can be decomposed as $\bar{\mathcal{A}} = [\mathbf{A}, \mathbf{B}]$ with $\mathbf{A} = \begin{bsmallmatrix}
    0   & \cdots & 0       & A_{T-1} \\
    A_0 &        &         & 0       \\
        & \ddots &         & \vdots  \\
    0   &        & A_{T-2} & 0
\end{bsmallmatrix} \in \mathbb{R}^{n_\psi T\times n_\psi T}$ and $\mathbf{B} = \begin{bsmallmatrix}
    0   & \cdots & 0       & B_{T-1} \\
    B_0 &        &         & 0       \\
        & \ddots &         & \vdots  \\
    0   &        & B_{T-2} & 0
\end{bsmallmatrix} \in \mathbb{R}^{n_\psi T\times n_uT}$. Thus, we can write phase-binned EDMDc as the following minimization:
\begin{equation}
\label{eq:edmd}
    \min_{A_{\sigma_k}, B_{\sigma_k}}\sum_i^D ||\psi(x^i_{k+1}) - A_{\sigma^i_k}\psi(x^i_k) - B_{\sigma^i_k}u^i_k||^2_2, 
\end{equation}
with $A_{\sigma_k+T} = A_{\sigma_k}$ and $B_{\sigma_k+T} = B_{\sigma_k}$. This implies that for a periodic system, phase-binned EDMDc solves EDMD for each phase $\sigma_k$, yielding a linear time-periodic representation. Choosing the first $n_x$ set of observables as the system state and reordering $\psi(\cdot)$ such that $\psi_l(x_k)  = x_k^{(l)}$ for all $l\in \{1, \dots, n_x\}$, we get the following system of linear equations.
\begin{equation}
\label{eq:ltp_predictor}
    z_{k+1} = A_{\sigma_k}z_k+B_{\sigma_k}u_k
\end{equation}
with $z_0 = [x_0^\top, \psi_1(x_0), \dots, \psi_{n_\psi -n_x}(x_0)]^\top$. By choosing $C = [I,0]$, where $I$ is the identity matrix of size $n_x$, we can retrieve the state with $x_k = Cz_k$. 
\subsection{Relationship between True and Lifted Monodromy Matrix}
\label{sec:monodromy_relation}
Similar to the monodromy matrix of the nonlinear system \eqref{eq:monodromy}, the predictor \eqref{eq:ltp_predictor} has its own $T$-step transition matrix, or \emph{lifted monodromy matrix}. We obtain it by composing the phase-indexed blocks over a full period as
\begin{equation}
\label{eq:lifted_monodromy}
    M_{\sigma_k}^z \coloneqq A_{\sigma_k+T-1}\cdots A_{\sigma_k} \in \mathbb{R}^{n_\psi \times n_\psi}, \quad \sigma_k \in \mathbb{Z}_T,
\end{equation}
Since $M^z_{\sigma_k}$ governs the response of the predictor over a period, it determines the periodic stability of \eqref{eq:ltp_predictor} over a period of $T$.

The lifted monodromy \eqref{eq:lifted_monodromy} acts on $\mathbb{R}^{n_\psi}$ while the true monodromy \eqref{eq:monodromy} acts on $\mathbb{R}^{n_x}$, so the two are not directly comparable. However, we can derive a relationship between them via the geometry of the dictionary along the orbit. Let
\begin{equation}
\label{eq:dictionary_jacobian}
    J_{\sigma_k} \coloneqq D\psi(\bar{x}_k) = \frac{\partial \psi}{\partial x}\bigg|_{(\bar{x}_k, \bar{u}_k, \sigma_k)} \in \mathbb{R}^{n_\psi \times n_x},
\end{equation}
which sends a tangent perturbation $\delta x$ at $\bar{x}_k$ to the induced perturbation $\delta\psi = J_{\sigma_k}\,\delta x$ of the lifted state. Since $\psi(\cdot)$ contains the state after the re-ordering as the first $n_x$ entries, $J_{\sigma_k} = [\,I_{n_x};\; \star\,]$ has full column rank for every $\sigma_k$, and $J_{\sigma_k+T} = J_{\sigma_k}$ by periodicity of the orbit.
\begin{assumption}[Local dictionary invariance]
\label{as:invariance}
The dictionary $\psi(\cdot)$ is invariant in a neighborhood of the orbit. In other words, for each $\sigma_k$, there exist $A_{\sigma_k} \in \mathbb{R}^{n_\psi \times n_\psi}$ and $B_{\sigma_k} \in \mathbb{R}^{n_\psi \times n_u}$ such that
\begin{equation}
\label{eq:invariance}
    \psi\big(f_p(x,u,\sigma_k)\big) = A_{\sigma_k}\,\psi(x) + B_{\sigma_k}\,u
\end{equation}
for all $x$ in an open neighborhood of $\bar{x}_k$ and all $u$ in an open neighborhood of $\bar{u}_k$. Moreover, $\psi(\cdot)$ is continuously differentiable on these neighborhoods.
\end{assumption}  

\begin{theorem}[Monodromy relation]
\label{thm:embedding}
Consider a nonlinear $T-$periodic system in \eqref{eq:nonlinear_dyn} with $p\in \mathcal{P}$ as the system parameters. Suppose $\{\bar{\mathbf{x}},\bar{\mathbf{u}}\}$ be a $T$-periodic orbit of \eqref{eq:nonlinear_dyn} and Assumption~\ref{as:invariance} hold. Then, the following statements hold:
\begin{enumerate}
    \item \emph{(Intertwining)} For every $\sigma_k \in \mathbb{Z}_T$,
    \begin{equation}
    \label{eq:intertwining}
        A_{\sigma_k} J_{\sigma_k} = J_{\sigma_{k+1}}\Omega_{\sigma_k}^p, \qquad B_{\sigma_k} = J_{\sigma_{k+1}}\Gamma_{\sigma_k}^p .
    \end{equation}
    \item \emph{(Embedding)} The lifted monodromy $M^z_0 \coloneqq A_{T-1}\cdots A_0$ and the true monodromy $M_0^x(p) \coloneqq \Omega_{T-1}^p\cdots\Omega^p_0$ satisfy
    \begin{equation}
    \label{eq:monodromy_embedding}
        M_0^z J_0 = J_0 M_0^x(p).
    \end{equation}
    \item \emph{(Spectral inclusion)} $\operatorname{range}(J_0)$ is an $M^z$-invariant subspace, and consequently
    \begin{equation}
    \label{eq:spectral_inclusion}
        \Lambda(p) = \operatorname{spec}(M^x(p)) \subseteq \operatorname{spec}(M^z).
    \end{equation}
\end{enumerate}
\end{theorem}

\begin{proof}
As a result of Assumption \ref{as:invariance}, both sides of \eqref{eq:invariance} agree on an open set, implying that their derivatives must also agree. Taking the partial derivative \eqref{eq:invariance} with respect to $x$ at $(\bar{x}_k,\bar{u}_k)$ and using the orbit relation $x_{k+1}=f_p(x_k,u
_k,\sigma_k)$, the chain rule gives on the left
\begin{align}
     D\psi\big(f_p(x_k,u_k,\sigma_k)\big)\, Df_p(x_k,u_k,\sigma_k) \\= D\psi(x_{k+1})\Omega^p_k = J_{\sigma_{k+1}}\Omega^p_{\sigma_k},
\end{align}
and on the right $A_{\sigma_k} D\psi(\bar{x}_k) = A_{\sigma_k} J_{\sigma_k}$. Similarly, taking the partial derivative with respect to $u$ gives $J_{\sigma_{k+1}}\Gamma^p_{\sigma_k}$ on the left and $B_{\sigma_k}$ on the right. This proves the first property.

We now apply \eqref{eq:intertwining} repeatedly and use the definition of the \emph{monodromy matrix} for the true and the lifted system:
\begin{equation}
\label{eq:telescope}
\begin{aligned}
    A_0 J_0 &= J_1 \Omega^p_0, \\
    A_1 A_0 J_0 &= A_1 J_1 \Omega^p_0 = J_2 \Omega^p_1 \Omega^p_0, \\
    &\;\;\vdots \\
    \underbrace{A_{T-1}\cdots A_0}_{M_0^z} J_0 &= J_T \underbrace{\Omega^p_{T-1}\cdots\Omega^p_0}_{M_0^x(p)} .
\end{aligned}
\end{equation}
Since $\bar{x}_T = \bar{x}_0$ and $J_{\sigma_k}$ is defined by evaluating $D\psi(\cdot)$ at the orbit point, $J_T = J_0$, which yields \eqref{eq:monodromy_embedding}, proving the second property. Note that $J_0$ has full column rank $n_x$ (by construction), where $J_{\sigma_k} = [\,I_{n_x};\; \star\,], \ \forall \sigma_k \in \mathbb{Z}_T$.

Let $\mathcal{J} \coloneqq \operatorname{range}(J_0) = \{J_0 v : v \in \mathbb{R}^{n_x}\} \subseteq \mathbb{R}^{n_\psi}$, a subspace of dimension $n_x$. We choose any $w \in \mathcal{J}$, so that $w = J_0 v$ for some $v \in \mathbb{R}^{n_x}$. Then
\begin{equation}
\label{eq:invariance_step}
    M^z w = M^z J_0 v = J_0\big(M^x(p)\, v\big),
\end{equation}
which is $J_0$ applied to the vector $M^x(p)v \in \mathbb{R}^{n_x}$, hence again an element of $\mathcal{J}$. Therefore $M^z \mathcal{J} \subseteq \mathcal{J}$.

We are now ready to show that $\mu \in \operatorname{spec}\big(M^x(p)\big)$ with eigenvector $v \neq 0$, i.e. $M^x(p)v = \mu v$, is also in $\operatorname{spec}\big(M^z\big)$. Let $w \coloneqq J_0 v$ be any vector. Then, we can write the following:
\begin{equation}
\label{eq:eigen_lift}
    M^z w = M^z J_0 v = J_0 M^x(p) v = J_0(\mu v) = \mu w .
\end{equation}
Since $v \neq 0 \implies w \neq 0$. Hence $w$ is an eigenvector of $M^z$ and $\mu\in \Lambda(p) \subseteq \operatorname{spec}(M^z)$, proving the third property. 
%This proves the third property.
\end{proof}
\begin{figure}[t]
\centering
\includegraphics[width=0.33\textwidth, trim = 0.6cm 8cm 0.6cm 3.0cm,clip]{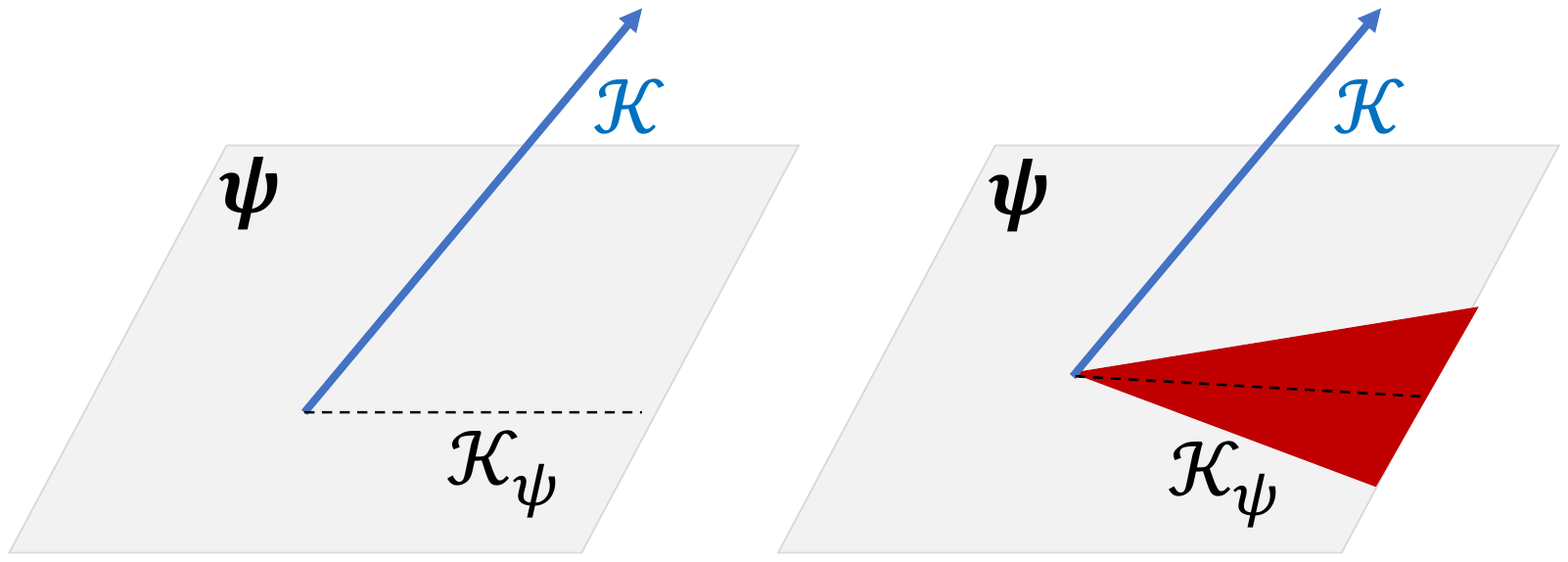}
\caption{Illustration of Floquet-parameterized EDMD (\textbf{right}) and purely EDMD (\textbf{left}). The main idea is to constrain the search space using \emph{Floquet multipliers} along a periodic orbit.}
\label{fig:floquet}
\end{figure}
\subsection{Floquet-Parameterized Koopman EDMDc}
\label{sec:constrained_edmd}

We now leverage Theorem~\ref{thm:embedding} to constrain the search space of system matrices and make the linear predictors physically consistent with the system's nonlinear dynamics around the periodic orbit. The periodic orbit yields $\{\bar{x}_k,\bar{u}_k\}_{\sigma_k\in\mathbb{Z}_T}$, and $J_{\sigma_k} = D\psi(\bar{x}_k)$ is available in closed form once the dictionary is fixed. However, the variational matrices $\Omega^p_{\sigma_k},\Gamma^p_{\sigma_k}$ depend on the model parameters $p\in \mathcal{P}$ not exactly known. Conceptually, the idea is to constrain the search space in the subspace defined by user-defined observable functions formed by the dictionary, as shown in Fig. \ref{fig:floquet}. 

We model this uncertainty with a phase-wise polytopic set. For each $\sigma_k\in\mathbb{Z}_T$,
\begin{equation}
    \Delta_{\sigma_k} = \mathrm{Co}\big\{[\Omega^{(1)}_{\sigma_k}\ \Gamma^{(1)}_{\sigma_k}],
    \dots,[\Omega^{(n_\lambda)}_{\sigma_k}\ \Gamma^{(n_\lambda)}_{\sigma_k}]\big\},
\end{equation}
where the vertices collect the variational matrices over the parametric variation $p \in \mathcal{P}$ and form a convex hull, so the vertex index refers to the same parameter at every phase.

\begin{assumption}
\label{ass:polytopic}
The vertices $\{[\Omega^{(s)}_{\sigma_k}\ \Gamma^{(s)}_{\sigma_k}]\}_{s=1}^{n_\lambda}$ are known for every $\sigma_k\in\mathbb{Z}_T$, obtained by solving the variational equations and retrieving a convex hull along the periodic orbit for each $p \in \mathcal{P}$.
\end{assumption}

Both Assumptions \ref{as:invariance} and \ref{ass:polytopic} are relatively mild and practical, as smooth observables are readily chosen in practice and convex hulls of the variational matrices can be easily constructed (e.g., parameter-affine systems).
We introduce a weighting parameter $\bm{\lambda} \in \mathbb{R}^{n_\lambda}:= \{ \forall s\in \{ 1, \dots, n_\lambda\}, \ \lambda_s \geq 0, \ \bm{1}^\top\bm{\lambda} = 1\}$ as a decision variable that weighs the contribution of each vertex in $\Delta_{\sigma_k}$ across all phases. The Floquet-parameterized constrained
least-squares EDMD is
\begin{subequations}
\label{eq:constrained_edmd}
\begin{align}
    \min_{\{A_{\sigma_k},B_{\sigma_k},\bm{\lambda} \}}\ & \sum_{i=1}^{D}
      \big\|\psi(x^i_{k+1}) - A_{\sigma^i_k}\psi(x^i_k) - B_{\sigma^i_k}u^i_k\big\|_2^2 \nonumber\\
    \text{s.t.}\quad
    & A_{\sigma_k} J_{\sigma_k} = J_{\sigma_{k+1}}
        \textstyle\sum_{s=1}^{N_v}\lambda_s\,\Omega^{(s)}_{\sigma_k}, \label{eq:const_state} \\
    & B_{\sigma_k} = J_{\sigma_{k+1}}
        \textstyle\sum_{s=1}^{N_v}\lambda_s\,\Gamma^{(s)}_{\sigma_k},\label{eq:const_inp}  \\
    & \forall s\in \{ 1, \dots, n_\lambda\}, \ \lambda_s \geq 0, \ \bm{1}^\top\bm{\lambda} = 1,\label{eq:const_conv}\\
    & \psi(\bar{x}_{k+1}) = A_{\sigma_k}\psi(\bar{x}_k) + B_{\sigma_k}\bar{u}_k, \label{eq:const_orbit} 
\end{align}    
\end{subequations}
with the first three constraints imposed for all $\sigma_k \in \mathbb{Z}_T$ and $J_T = J_0$ closing the loop. Note that the constraints are jointly affine in $(A_{\sigma_k},B_{\sigma_k},\bm{\lambda})$, thus \eqref{eq:constrained_edmd} is a quadratic program. The constraints in \eqref{eq:const_state}, \eqref{eq:const_inp} and \eqref{eq:const_conv} constrain the variational matrices of the lifted system along the reference orbit using the convex hull of the variational matrices of the original system. \eqref{eq:const_orbit} ensures that the lifted system satisfies the periodic orbit. Together, the constrained least-squares optimization ensures the lifted system captures the periodic stability properties of the original system.  
\subsection{Koopman-MPC for Tracking Periodic Orbits}
\label{sec:mpc}
We will now utilize the linear predictor obtained by solving the constrained least squares in \eqref{eq:constrained_edmd}. Note that unlike time-invariant regulation in \cite{korda2018linear}, for periodic tracking the decision variables are deviations from $\{\bar{\mathbf{x}},\bar{\mathbf{u}}\}$ rather than the absolute states, and the propagation matrices depend on the phase at which the problem is posed.

Consider $\psi(\bar{x}_k)$ as the lift of the reference orbit, which is $T$-periodic and computed offline. Define the lifted tracking error and input deviation
\begin{equation}
\label{eq:error_coords}
    e^z_k \coloneqq \psi(x_k) - \psi(\bar{x}_k), \quad v_k \coloneqq u_k - \bar{u}_k .
\end{equation}
Subtracting the reference evolution from \eqref{eq:ltp_predictor} gives
\begin{equation}
\label{eq:error_dyn}
    e^z_{k+1} = A_{\sigma_k} e^z_k + B_{\sigma_k} v_k.
\end{equation}
Let the tracked state error be defined as $e_k^x \coloneqq x_k - \bar{x}_k = C e^z_k$, since $C\psi(x_k) = x_k$ by construction of the dictionary.

Let the current step be denoted by $k$ with phase $\sigma_k$, and let $H$ be the prediction horizon. We can write the transition matrix  of \eqref{eq:error_dyn} for all $i>j$ as follows:
\begin{equation}
\label{eq:transition}
    \Theta_{\sigma_k}^{(i,j)} \coloneqq A_{\sigma_k+i-1}\cdots A_{\sigma_k+j}.
\end{equation}
Recursively eliminating the lifted errors from \eqref{eq:error_dyn} and using the relation $C\psi(x_k) = x_k$ yield
\begin{equation}
\label{eq:condensed}
    \begin{bsmallmatrix} e^x_{k+1} \\ \vdots \\ e^x_{k+H} \end{bsmallmatrix}
    = \mathbf{E}_{\sigma_k} e^z_k + \mathbf{F}_{\sigma_k} \begin{bsmallmatrix} v_k \\ \vdots \\ v_{k+H-1}\end{bsmallmatrix},
\end{equation}
where 

$
    \mathbf{E}_{\sigma_k} \coloneqq \begin{bsmallmatrix} C\Theta_{\sigma_k}^{(1,0)} & C\Theta_{\sigma_k}^{(2,0)} & \dots & C\Theta_{\sigma_k}^{(H,0)} \end{bsmallmatrix}^\top \in \mathbb{R}^{n_xH}, \
    \mathbf{F}_{\sigma_k} \coloneqq \begin{bsmallmatrix} CB_{\sigma_k} & \dots & 0 \\ C\Theta_{\sigma_k}^{(1,0)}B_{\sigma_k} & CB_{\sigma_k}&0\\ \vdots & \vdots & \vdots \\ C\Theta_{\sigma_k}^{(H-1,0)}B_{\sigma_k} & C\Theta_{\sigma_k}^{(H-2,0)}B_{\sigma_k}& CB_{\sigma_k}\end{bsmallmatrix} \in \mathbb{R}^{n_xH \times n_uH}.
$

As the phase $\sigma_k$ is restricted to $\mathbb{Z}_T$, we have exactly $T$ distinct pairs $(\mathbf{E}_{\sigma_k}, \mathbf{F}_{\sigma_k})$. In practice, we assemble these offline and select them by table lookup at runtime. 

We are now ready to present the condensed form of the quadratic program for tracking MPC as follows:  
\begin{subequations}
\label{eq:qp}
\begin{align}
    \min_{V} \ & V^\top \bar{R} V + \big(\mathbf{E}_{\sigma_k} e^z_k + \mathbf{F}_{\sigma_k} V\big)^\top \bar{Q}\, \big(\mathbf{E}_{\sigma_k} e^z_k + \mathbf{F}_{\sigma_k} V\big) \nonumber\\
    \text{s.t.} \quad &  \mathbf{E}_{\sigma_k} e^z_k + \mathbf{F}_{\sigma_k} V \in \mathcal{E}^x_{\sigma_k}, \label{eq:qp_state}\\
    &  V \in \mathcal{E}^u_{\sigma_k}, \label{eq:qp_input}
\end{align}
\end{subequations}
where $V = \begin{bmatrix} v_k^\top & \dots & v_{k+H-1}^\top\end{bmatrix}^\top \in \mathbb{R}^{n_uH}$, $\bar{R} = \operatorname{blkdiag}(R,\dots,R)$ and $\bar{Q} = \operatorname{blkdiag}(Q,\dots,Q,P_{\sigma_k+H})$.
$\mathcal{E}^u_{\sigma_k}$ and $\mathcal{E}^u_{\sigma_k}$ are the error sets in state and input, i.e., the admissible deviations from the reference orbit, defined as
\begin{equation}
\label{eq:error_sets}
     \forall \sigma_k \in \mathbb{Z}_T, \quad \mathcal{E}^x_{\sigma_k} \coloneqq \mathcal{X} - \bar{x}_{\sigma_k}, \qquad \mathcal{E}^u_{\sigma_k} \coloneqq \mathcal{U} - \bar{u}_{\sigma_k}. \nonumber
\end{equation}
The terminal weight $P_{\sigma_k}$ is the $T$-periodic solution of the discrete periodic Riccati equation associated with $\{A_{\sigma_k}, B_{\sigma_k}, Q, R\}$, and is also tabulated over the phases $\sigma_k$. The first block of the minimizer is applied, $u_k = \bar{u}_k + V^\star_{1:n_u}$, and the horizon recedes, and the process repeats.%we repeat the same process again for the next step.

% \begin{remark}[Cost of the lifting]
% \label{rem:qp_dimension}
% The decision vector $V$ has dimension $Hn_u$, independent of the lifted dimension $n_\psi$: the lifting inflates the offline matrices $E_\sigma, F_\sigma$ but not the online problem. The only run-time appearance of $n_\psi$ is the single evaluation $\psi(x_t)$ forming $e_t$, performed once per solve and not inside the solver iterations. A rich dictionary is therefore essentially free at run time, which is what makes the accuracy gains of Section~\ref{sec:constrained_edmd} worth pursuing.
% \end{remark}

%% file: sections/5_numerical_example.tex
\begin{figure}[t]
\centering
\includegraphics[width=0.32\textwidth,clip]{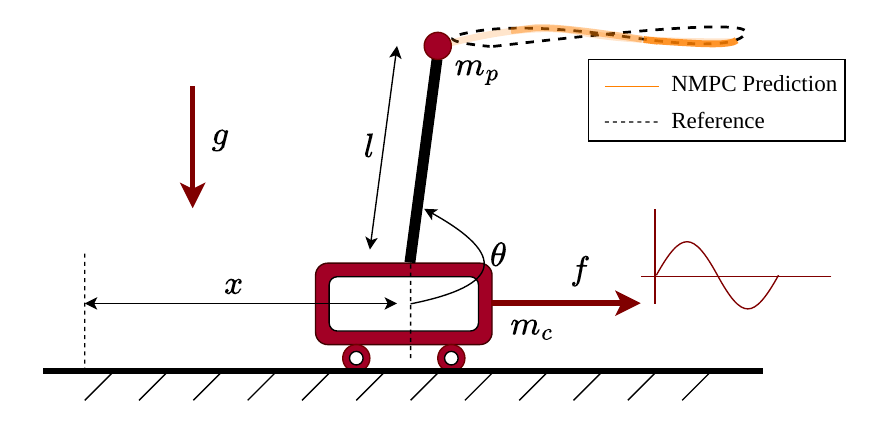}
\caption{A cartpole system under a forcing periodic input.}
\label{fig:cartpole}
\end{figure}

In this section, we show the effectiveness of our proposed framework for tracking a reference periodic orbit and demonstrate that it requires fewer training samples than purely data-driven EDMD-based Koopman-MPC \cite{korda2018linear} to achieve the same performance. We consider a cartpole system where the controller's task is to control a cart to track a periodic reference while ensuring the pole stands upright as shown in Fig.~\ref{fig:cartpole}. The continuous-time dynamics of a cartpole system \cite{tedrake2023underactuated} is as follows: 
\begin{multline}
\label{eq:cartpole}
\begin{bsmallmatrix} m_c + m_p & m_p \ell \cos\theta \\
                m_p \ell \cos\theta & m_p \ell^2 \end{bsmallmatrix}
\begin{bsmallmatrix} \ddot{x} \\ \ddot{\theta} \end{bsmallmatrix}
+ \begin{bsmallmatrix} -m_p \ell \dot\theta^2 \sin\theta \\ 0 \end{bsmallmatrix}
= \\ \begin{bsmallmatrix} 0 \\ -m_p g \ell \sin\theta \end{bsmallmatrix}
+ \begin{bsmallmatrix} 1 \\ 0 \end{bsmallmatrix} f,
\end{multline}
with state $s = [x, \theta, \dot{x}, \dot\theta]^\top$, input $u = f$, and $\theta$ measured from the hanging configuration, so that the inverted equilibrium is $\theta = \pi$. We consider that the nominal model has $m_c = 1.0$~kg, $m_p = 0.2$~kg, $\ell = 0.5$~m, $|f| \leq 20$~N, and discretize it with a sampling period of $\Delta t = 0.05$~s. 
The actual model differs with $m_p = 0.25$~kg, $\ell = 0.56$~m.
\begin{figure*}[t]
\centering
\includegraphics[width=\textwidth,clip]{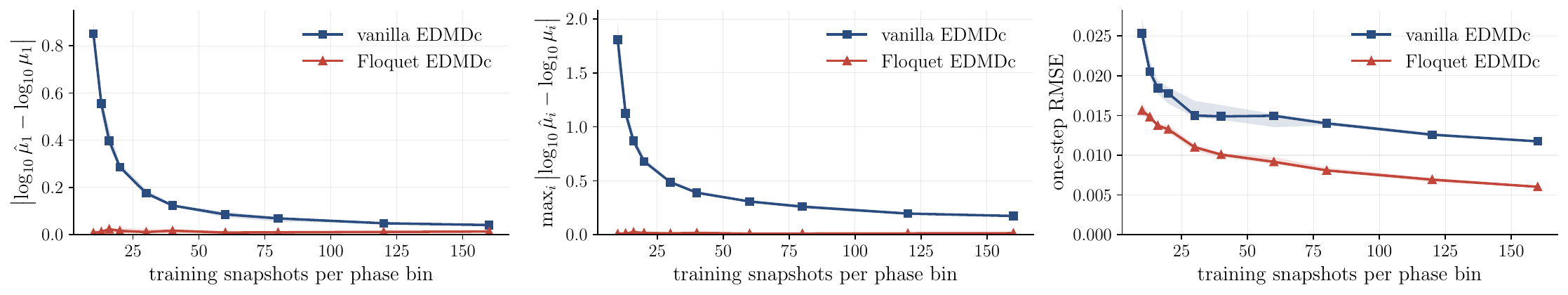}
\caption{Comparison between Vanilla and the proposed Floquet-parameterized EDMDc. \textbf{Left:} Difference between the dominant Floquet multiplier of the learned system and the true system, \textbf{Center:} Worst-of-all difference between the true and estimated Floquet multipliers, and \textbf{Right:} One-step root-mean-squared prediction error between linear predictor and true model.}
\label{fig:comparison}
\end{figure*}

The reference commands a $T$-periodic cart sweep $\bar{x}_k = a_1\sin(\frac{2\pi k\Delta t}{T}) + a_2 \sin(\frac{4 \pi k\Delta t}{T})$ with $T = 2.5$~s, $a_1 = 0.45$~m, $a_2 = 0.08$~m. We use a tracking nonlinear MPC to track this reference, thus obtaining a $T-$periodic reference orbit. The resulting monodromy matrix of the system along the reference orbit has Floquet multipliers $\Lambda = \{\,2.2285\times 10^{5},\; 1,\; 1,\; 4.4872\times 10^{-6}\,\}$. We collect the state-input trajectory data by random state perturbations and additive input dither on top of the closed-loop NMPC. We collect in total $D = 8000$ snapshot triples $(x^i_k, u^i_k, x^i_{k+1})$ tagged by orbit phase, i.e.\ $160$ per phase bin. As mentioned earlier, a model mismatch exists between the actual plant and the nominal model. We assume that we know the range of true $m_p$ and $\ell$, and thus we construct a convex set with $n_\lambda = 4$ vertices at the corners of $m_p \in [0.15, 0.30] \times \ell \in [0.44, 0.60]$ to obtain the variational matrices required in the constrained problem defined in \eqref{eq:constrained_edmd}. 

For solving the EDMDc (both \eqref{eq:edmd} and \eqref{eq:constrained_edmd}), we use $n_z = 20$ observables: the state itself, so that it is recoverable exactly from the lift, together with $16-$dimensional Gaussian radial basis function (RBF) kernels \cite{mauroy2020koopman} centered around the reference trajectory. We first evaluate the performance of our proposed method in preserving the Floquet multipliers (spectral accuracy of the periodic orbit) while varying the size of the training data. We compute and plot the errors in the estimated Floquet multipliers compared to the true multipliers. Since the lifted model has more multipliers, we use the Hungarian algorithm \cite{kuhn1955hungarian} with a logarithmic cost defined by $|\log_{10}(\hat{\mu}) - \log_{10}(\mu)|$ for corresponding assignments. The left and middle panels of Fig. \ref{fig:comparison} compare errors in the dominant Floquet multiplier and the worst-case estimation error among all Floquet multipliers. It clearly depicts that the proposed method learns a model that is consistent with the true Floquet multiplier using as few as $10$ samples per phase bin. We then compare the $1-$step open-loop root mean squared error (RMSE) of the two models. The rightmost panel of Fig. \ref{fig:comparison} shows that the proposed method, due to periodic orbital stability constraints, performs better than the baseline. Finally, we compare the closed-loop performance of both the vanilla EDMDc and Floquet EDMDc-based linear MPC methods. The tables show the closed-loop reference tracking error (error in the cart position from the reference trajectory) for both methods. We can clearly observe that the Floquet method learns a physically consistent model from a few snapshots of data compared to the vanilla method. Note that all experiments demonstrate one key insight: with more data, the performance of both methods becomes comparable, and the main benefit of our proposed approach is when there is a limited training data.

\begin{table}[t]
\centering
\caption{Median Root Mean Square Error (RMSE) of the cart position (in mm) for 100 Monte Carlo trials with perturbed initial conditions, comparing both methods trained using various training data sizes per phase bin. }
\label{tab:mismatch}
\begin{tabular}{lccccc}
\hline
\textbf{Method/Snapshots Per Phase}& 10 & 20& 40 & 80 &160\\
\hline
Vanilla EDMDc \cite{korda2018linear} & 1507.7 & 77.6 & 74.1& 73.4& 72.9\\
Floquet EDMDc (\textbf{Proposed}) & \textbf{73.7} & \textbf{73.3} & \textbf{73.1} & \textbf{71.9}& \textbf{71.6}\\
\hline
\end{tabular}
\end{table}

%% file: sections/6_conclusion.tex
% In this paper, we presented a constrained EDMD-based Koopman-MPC formulation for tracking periodic reference orbits. We leverage Floquet theory, which studies the stability of periodic systems, and derive a relationship between the monodromy matrices of the original and the lifted system. We showed that this relation can subsequently be used to formulate a constrained least-squares problem to learn a physically consistent model of the original periodic system. Furthermore, we leveraged the learned model and formulated a condensed quadratic program for MPC optimization. Finally, we validate the proposed method using numerical simulations of a cartpole system, where the system tracks a periodic orbit, and demonstrate its superior tracking performance over purely data-driven Koopman-MPC with fewer training samples. 
In this paper, we presented a constrained EDMD-based Koopman-MPC framework for periodic-orbit tracking that incorporates Floquet stability information through a relationship between the original and lifted-system monodromy matrices. We utilized this relation to formulate a constrained least-squares problem to learn a physically consistent model that can be subsequently used for MPC.
%This relationship enabled us to learn a physically consistent model via a constrained EDMD formulation, which is then used to formulate a condensed quadratic MPC problem. 
Finally, we validate the proposed method using numerical simulations of a cartpole system, where the cart tracks a periodic orbit while holding the pole upright, and demonstrate its superior tracking performance over purely data-driven Koopman-MPC with fewer training samples. 
%Numerical simulations on a periodically forced cart-pole demonstrate improved tracking performance over purely data-driven Koopman-MPC while requiring fewer training samples.